\documentclass{article}
\usepackage{spconf,amsmath,graphicx,pdfpages,hyperref,multirow,color,amssymb}

\usepackage{amsmath,amsfonts,bm}

\def\eqref#1{equation~\ref{#1}}

\def\1{\bm{1}}

\def\ve{{\bm{e}}}

\def\vh{{\bm{h}}}

\def\vp{{\bm{p}}}

\def\mE{{\bm{E}}}

\def\mH{{\bm{H}}}

\def\mK{{\bm{K}}}

\def\mP{{\bm{P}}}
\def\mQ{{\bm{Q}}}

\DeclareMathAlphabet{\mathsfit}{\encodingdefault}{\sfdefault}{m}{sl}
\SetMathAlphabet{\mathsfit}{bold}{\encodingdefault}{\sfdefault}{bx}{n}

\def\gC{{\mathcal{C}}}

\def\gE{{\mathcal{E}}}

\def\gL{{\mathcal{L}}}

\def\gS{{\mathcal{S}}}

\newcommand{\softmax}{\mathrm{softmax}}

\DeclareMathOperator*{\ptm}{PLM}
\DeclareMathOperator*{\mlp}{MLP}
\DeclareMathOperator*{\sep}{[SEP]}
\DeclareMathOperator*{\cls}{[CLS]}
\DeclareMathOperator*{\bmtwo}{BM25}
\DeclareMathOperator*{\skg}{SKG}
\DeclareMathOperator*{\pool}{MeanPooling}
\DeclareMathOperator*{\sattn}{Self-Attention}

\title{Unsupervised Knowledge Graph Construction and Event-centric Knowledge Infusion for Scientific NLI}

\name{Chenglin Wang$^1$, Yucheng Zhou$^2$, Guodong Long$^2$, Xiaodong Wang$^1$, Xiaowei Xu$^{1*}$\thanks{*Corresponding author. This work is supported by the National Key R\&D Program of China (No. 2020YFB1710005).}}
\address{$^1$School of Computer Science
and Technologys, Ocean University of China\\ 
$^2$Australian AI Institute, School of Computer Science, FEIT, University of Technology Sydney}

\begin{document}
\maketitle
\begin{abstract}
With the advance of natural language inference (NLI), a rising demand for NLI is to handle scientific texts. Existing methods depend on pre-trained models (PTM) which lack domain-specific knowledge. To tackle this drawback, we introduce a scientific knowledge graph to generalize PTM to scientific domain. However, existing knowledge graph construction approaches suffer from some drawbacks, i.e., expensive labeled data, failure to apply in other domains, long inference time and difficulty extending to large corpora. Therefore, we propose an unsupervised knowledge graph construction method to build a scientific knowledge graph (SKG) without any labeled data. Moreover, to alleviate noise effect from SKG and complement knowledge in sentences better, we propose an event-centric knowledge infusion method to integrate external knowledge into each event that is a fine-grained semantic unit in sentences. Experimental results show that our method achieves state-of-the-art performance and the effectiveness and reliability of SKG.
\end{abstract}


\section{Introduction}
\label{sec:intro}
Natural language inference (NLI), an essential task for natural language understanding, aims to deduce relationship between the given premise and hypothesis \cite{DBLP:conf/emnlp/BowmanAPM15}. NLI is a fundamental problem in many natural language processing (NLP) tasks, such as sentence embeddings \cite{DBLP:conf/emnlp/GaoYC21}, question answering \cite{TanSXZ16QALSTM} and commonsense reasoning \cite{DBLP:conf/acl/SapSBCR20commonsense}. Therefore, it has been widely concerned by many researchers. With the widespread NLP application, a rising demand for NLI methods is to handle specific-domain text, such as scientific texts \cite{SCiNLI}, medical articles \cite{HeZZCC20} and financial news \cite{SchumakerC09financial}. To build a large NLI dataset related to scientific texts, SciNLI dataset \cite{SCiNLI} is built from scholarly papers on NLP and computational linguistics. 

Due to the success of pre-trained language models (PTM) (e.g., BERT \cite{BERT} and RoBERTa \cite{RoBERTa}), a general paradigm for NLI is to fine-tune a PTM on downstream dataset. However, PTMs fine-tuned on specific-domain data often suffer from a cross-domain problem since they are pre-trained on general domain corpora such as news articles and Wikipedia. Many works generalize PTM to specific-domain via pre-train on specific-domain corpus \cite{SciBERT,Zhou22Sketch} or introducing specific-domain knowledge graph \cite{zhou2021modeling}. Although Beltagy \textit{et al.} \cite{SciBERT} consume enormous training resources to exclusively pre-train SciBERT on scientific texts, RoBERTa with a more sophisticated pre-training leads to better performance. Therefore, it is necessary to introduce a scientific knowledge graph to generalize PTM to scientific domain.

Recently, some works \cite{SemEval,SCIERC} propose building scientific knowledge graphs automatically via training entity recognition and relationship extraction models with labeled data. Despite their success, they still suffer from some drawbacks, i.e., expensive labeled data, failure to apply in other domains, long inference time \footnote{In our pilot experiments on 8-core-CPU/Nvidia RTX2080 GPU, dependency-parser from Stanford-stanza processed 833 sentences/second, which is 110x faster than S{\scriptsize CI}IE \cite{SCIERC}.} and difficulty extending to large corpora. Specifically, existing methods \cite{SemEval,SCIERC} are trained on only 500 abstracts and show an undesirable performance (i.e., 44.7 on F1). Therefore, we efficiently and easily build a scientific knowledge graph (SKG) without any labeled data. We first parse scientific texts via a dependency parser and then extract the subjects, predicates and objects as triplet candidates. To reduce noise samples in triplet candidates, we propose heuristic filtering methods to improve the accuracy of triplets. Due to fast inference and not requiring any labeled data and training sources, our method can easy to extend to large corpora in other domains.

To improve PTM reasoning, previous works \cite{WangLZ20KGNLI} mainly use sentences as queries to retrieve triplets in knowledge graph and integrate them into PTM. However, since there are still some noise samples in SKG, directly integrating knowledge into model damages model learning. Recently, event-centric reasoning \cite{Zhou22EventBERT,Zhou22ClarET} shows powerful reasoning capability in context via understanding correlation between events. To reduce the effect of noise data and complement knowledge in sentences better, we propose an event-centric knowledge infusion framework. Precisely, we follow \cite{Zhou22EventBERT} to split events into sentences and then use all events as queries to retrieve relevant triplets, which can prevent the retrieved knowledge only relevant to limited semantics in sentences. Moreover, we integrate knowledge into multiple event units, improving context reasoning via enriching semantic information in each event.

We conduct an extensive evaluation on SciNLI dataset \cite{SCiNLI}. Results show that our method achieves state-of-the-art performance compared with other methods. In addition, we analyze the effectiveness and reliability of SKG.

\section{Unsupervised SKG Construction}
Existing methods \cite{SemEval,SCIERC} adopt entity and relation extraction paradigms to build scientific knowledge graphs automatically. However, these methods demand expensive labeled data and large training sources for training entity and relation extraction models. Moreover, since existing datasets for scientific knowledge graph construction are only labeled on abstract in NLP papers \cite{SemEval,SCIERC}, trained models are difficult to extend to large corpora and fail to apply in other domains. Therefore, we propose an efficient and easy knowledge graph construction method without any labeled data. Although we employ the method to build a scientific knowledge graph (SKG), it is easy to apply in other domains because it is domain-agnostic.

Specifically, we first leverage the dependency parser of Stanford CoreNLP \cite{corenlp} to parse all sentences in scientific corpus. Then, we locate predicates based on parsing results and extract related subject and object chunks. We collect a triplet set with many triplets consisting of subject chunks, predicates and object chunks. Although the triplet extraction is fast, there are still some wrong samples in the triplet set. To improve the accuracy of triplet set, we take three measures to filter and calibrate the triplet set. One is to filter words without specific meaning, such as letters and numbers. The second is to remove stopwords in triplets. The last is to filter low-frequency entities via a threshold $\lambda$ because a correct concept is usually more widely used. Since the whole ACL anthology is enormous and requires more computing resources, we build the SKG on the SciNLI training set with $\lambda$ of 1, which is a subset of ACL anthology. However, our method can be easily extended to entire ACL anthology. We will release a large version of SKG extracted from entire ACL anthology later.

\section{Method}
\label{sec:method}
This section start with a base SciNLI model. Then, details of event-centric knowledge infusion method are elaborated. 

\subsection{Base SciNLI Model}
Given a sentence pair $(\mP=\{p_{1},p_{2}...p_{m}\},\mQ=\{q_{1},q_{2}...q_{n}\})$, SCiNLI aims to recognize their semantic relationship $y$ (e.g., contrasting, reasoning, entailment and neutral). Due to the success of pre-trained language models (PTM), a general paradigm for natural language processing (NLP) tasks is to fine-tune a PTM on downstream dataset. In this work, our base SciNLI model is PTM, followed by a multilayer perceptron (MLP) with softmax. Specifically, we first concatenate the given sentence pair via a segment token $\sep$ and then pass them into the PTM followed by MLP, i.e.,
\begin{align}
\vh, \mH &= \ptm(\cls \mP \sep \mQ) \notag \\
&\text{where,}~ \mH \in \{\vh_1, \vh_2, \cdots, \vh_l\} \label{equ:hidden}\\
\vp &= \softmax(\mlp(\vh))  \label{equ:mlp}
\end{align}
where $\vh$ denotes the representation of sentence pair; $\vh_l$ represents token representation, and $l$ is length of input sequence. $\vp$ denote a probability distribution over four class $\gC$. 
Lastly, we train the base SciNLI model via maximum likelihood estimation, and the training loss function is defined as:
\begin{align}
    \gL = - \dfrac{1}{|\gS|}\sum_{\gS}\log\vp_{[y=c]}, \label{equ:loss_cls}
\end{align}
where $\gS$ denotes the whole training set; and $c$ refers to a ground truth class, and $c \in \gC$.

\subsection{Event-centric Knowledge Infusion}
However, due to the cross-domain gap between pre-trained corpus and SciNLI, fine-tuned PTM fails to handle scientific texts with complex logic and reasoning effectively. Moreover, PTM pre-trained on scientific texts via enormous training resources also underperforms. Therefore, we propose an event-centric knowledge infusion (EKI) method to integrate SKG into PTM. Our method contains event segment and retrieval, knowledge infusion and joint reasoning.

\noindent \textbf{Event Segment and Retrieval.} 
Previous works conduct retrieval via sentence as query and integrate retrieved results into PTM \cite{WangLZ20KGNLI}. However, these methods suffer from two drawbacks: First, retrieved results are only relevant to some sentence semantics. Secondly, incomprehensive retrieval results fail to effectively provide external knowledge required for PTM. To comprehensively introduce external knowledge to PTM, we employ fine-grained semantic units (i.e., events) as queries. Moreover, due to SKG build in an unsupervised manner, it still has some noise samples. Since semantic information in events is clear and lite, integrating knowledge into event can effectively alleviate effect of wrong information. Following \cite{Zhou22EventBERT}, we first parse a sentence via dependency parsing and then split a sentence into multiple events via connecting verb and relevant word chunks based on parsing results. Given a sentence pair $(\mP ,\mQ)$, extracted events denote as $\gE = \{\mE_1,\cdots,\mE_h\}, \forall \mE_i \in \mP \cup \mQ$. Next, we use $\bmtwo$ to retrieve top-$k$ triplets in SKG via events as queries, i.e., 
\begin{align}
\mK_i = \bmtwo(\mE_i, \skg), \mK= \{\mK_1,\cdots,\mK_h\}
\end{align}
where $\mK_i$ denotes top-$k$ retrieved triplet set for event $\mE_i$.

\noindent \textbf{Knowledge Infusion.} 
To integrate knowledge $\mK$ into PTM, we first concatenate triplets in $\mK_i$ and then pass them to PTM, i.e.,
\begin{align}
\vh^{(k)}_i\!\!=\!\!\ptm(\cls \mK_i^{(1)} \sep \cdots \sep \mK_i^{(k)} \sep)
\end{align}
where $\vh^{(k)}_i$ denote representation of triplet set $\mK_i$, and $\hat \mK = \{\vh^{(k)}_1, \vh^{(k)}_2, \cdots, \vh^{(k)}_h\}$. To obtain event representations of sentence pair, we conduct mean pooling operation on token representations $\mH$ from Equ.\ref{equ:hidden} based on extracted events $\gE$, i.e.,
\begin{align}
\ve_i = \pool(\vh_{st}, \vh_{end}), \{st, end\} \in \mE_i
\end{align}
where $\ve_i$ is event representation, and $\hat \mE = \{\ve_1, \ve_2, \cdots, \ve_h\}$. Lastly, we concatenate $\hat \mK$ and $\hat \mE$ to obtain knowledge-augmented events $\tilde \mE = \{\tilde \ve_1, \tilde \ve_2, \cdots, \tilde \ve_h\}$

\noindent \textbf{Joint Reasoning.} 
To jointly reason combined events and external knowledge better, we first use a self-attention module \cite{Vaswani17Attention} to deduce the relation between each knowledge-augmented event. Moreover, attention-based methods can learn adaptively weight for each representation to alleviate the effect of noise triplets in SKG, i.e.,
\begin{align}
\vh^{(\mu)}_i &= {\mlp}_\mu(\tilde \ve_i), \ve_i \in \tilde \mE, \mu \in \{Q,K,V\} \notag \\
&\text{where,}~ \mH^{(\mu)} \in \{\vh^{(\mu)}_1, \vh^{(\mu)}_2, \cdots, \vh^{(\mu)}_h\} \\
\tilde \mH &= \sattn(\mH^Q, \mH^K, \mH^V)   \\
\tilde \vh &= \pool(\tilde \mH)
\end{align}
where $\tilde \vh$ denotes a knowledge-augmented representation. Next we concatenate it and $\vh$ in Equ \ref{equ:hidden}, and pass to a MLP follow $\softmax$, i.e.,
\begin{align}
\vp^* &= \softmax(\mlp([\vh;\tilde \vh]))  
\end{align}
where $[;]$ denotes concatenate operation. $\vp^*$ denote a probability distribution over four class $\gC$. 

During model training, we train our model via maximum likelihood estimation, i.e.,
\begin{align}
    \gL^* = - \dfrac{1}{|\gS|}\sum_{\gS}\log\vp_{[y=c]}
\end{align}
where $\gS$ denotes the whole training set; and $c$ refers to a ground truth class, and $c \in \gC$.

\section{Experiments}
\label{sec:exp}
\subsection{Dataset and Evaluation Metrics}
We evaluate our proposed approach on the SciNLI dataset collected by \cite{SCiNLI}. SciNLI focuses on the semantic relations that are either relevant to the task of NLI or highly frequent in scientific text and leverages linking phrases to create the first-ever scientific NLI dataset. The dataset is a natural language inference dataset based on scientific text, which contains about 110k sample pairs and is divided into four output classes: contrasting, reasoning, entailment and neural. To equally distribute across the classes, the number of each class is the same when dividing train, dev and test, which are 25,353, 1,000 and 2,000, respectively. In addition, we use two different settings on the dataset to build our scientific knowledge graph. The first is to use only the training set to evaluate our method's effectiveness. The other is to use training, dev and test sets to analyze the performance of our unsupervised knowledge graph construction method. Moreover, we adopt two official evaluation metrics, accuracy (ACC) and F1-score(F1), to measure performance.

\begin{table}[t]\small
\centering
\setlength\tabcolsep{5.6pt}
\begin{tabular}{lcccccc}
\hline
\bf Method
& \multicolumn{1}{c}{\bf C} &  \multicolumn{1}{c}{\bf R} &  \multicolumn{1}{c}{\bf E}
&  \multicolumn{1}{c}{\bf N}&  \multicolumn{1}{c}{\bf F1}
&  \multicolumn{1}{c}{\bf ACC}\\ \hline
Lexicalized& 50.28& 37.18& 44.82& 55.77& 47.01& 47.78\\
CBOW& 54.62& 50.54& 52.33& 49.25& 51.68& 51.78\\
CNN& 63.73& 58.86& 62.66& 56.40& 60.41& 60.53\\
BiLSTM& 63.93& 57.32& 64.01& 59.25& 61.12& 61.32\\
BERT& 77.46& 71.74& 75.09& 76.47& 75.19& 75.17\\
SciBERT& 80.30& 74.18& 75.90& 79.76& 77.53& 77.52\\
RoBERTa& 81.18& 74.22& 77.99& 78.86& 78.06& 78.12\\
XLNet& 81.53& 75.95& 77.63& 77.63& 78.18& 78.23\\
EKI (Ours) & 81.76& 76.25& 78.66& {\textbf{80.02}} & 79.17& 79.20\\
EKI* (Ours) & {\textbf{82.51}} & {\textbf{77.38}} & {\textbf{78.86}} & 78.82& {\textbf{79.39}} & {\textbf{79.43}} \\ \hline
\end{tabular}
\caption{Comparison results on SciNLI test set.  C, R, E and N are the abbreviations of contrasting, reasoning, entailment and neural, respectively. * denotes that SKG used for our method is built on training, dev and test sets.}
\label{tab1}
\end{table}

\subsection{Experimental Setting}
The pre-train language model we used is the RoBERTa-base model \cite{RoBERTa}. The embedding size and hidden size in the model are set to 768. The num of heads in self-attention is set to 2, with the dropout set to 0.3. The maximum event number for the input sentence pair is 10, and the maximum length of the sentence pair is set to 196.
The number of retrieved triplets $k$ of each event is also set to 10, and the maximum length of these triplets after concatenation is set to 50. 
For model training, we use AdamW as our optimizer to optimize the cross-entropy loss with a learning rate of 5e-5.
The weight decay and linear warm-up step are 1.0 and 1,000.
We employ the ReLU activation function in all feed-forward networks in MLP. 
The maximum training epoch and batch size are set to 5 and 64, and a patience size of 2 about early stopping. 

\subsection{Main Results}
We evaluate our method on SciNLI, and experimental results are shown in Table \ref{tab1}. Comparison methods include three types of models: traditional machine learning models (e.g., lexicalized classifier), deep neural network models (e.g., BiLSTM, CBOW and CNN) and PTM (e.g., BERT, SciBERT, RoBERTa and XLNet). Results show that our approach achieves state-of-the-art performance. Our method outperforms RoBERTa, which shows the effectiveness of event-centric knowledge infusion of scientific text. In addition, to investigate the potential of our unsupervised scientific knowledge graph (SKG) construction methods, we evaluate our method with SKG built on training, dev and test set. Results show that performance of our method can improve again. The reason is that test set includes some concepts that do not appear in training set, and SKG built on texts on test set can alleviate this information gap. It shows that event-centric infusing knowledge is effective in scientific texts.

\subsection{Ablation Study}
\begin{table}[t]\small
\centering
\begin{tabular}{lcccccc}
\hline
\bf Method
& \multicolumn{1}{c}{\bf C} &  \multicolumn{1}{c}{\bf R} &  \multicolumn{1}{c}{\bf E}
&  \multicolumn{1}{c}{\bf N}&  \multicolumn{1}{c}{\bf F1}
&  \multicolumn{1}{c}{\bf ACC}\\ \hline
EKI & {\textbf{81.76}}  & {\textbf{76.25}} & {\textbf{78.66}} & {\textbf{80.02}} & {\textbf{79.17}} & {\textbf{79.20}} \\
w/ CLS & 81.59& 75.84& 77.94& 78.66& 78.51& 78.53\\
w/ Sent & 81.36& 74.88& 78.09& 78.73& 78.26& 78.33\\
w/o EKI & 81.18& 74.22& 77.99& 78.86& 78.06& 78.12\\\hline
\end{tabular}
\caption{\small Results of ablation Study. \textit{w/ CLS} denotes integrating external knowledge via concatenating CLS representations of the external knowledge and sentences. \textit{w/ Sent} denotes integrating external knowledge via concatenating the external knowledge with the sentence directly. \textit{w/o EKI} denotes removing our methods and external knowledge.}
\label{tab2}
\end{table}
Results of the ablation study are shown in Table \ref{tab2}. Firstly, our method outperforms w/ CLS, which demonstrates the effectiveness of event-centric knowledge infusion. w/ Sent also has a performance drop compared with our method. These demonstrate that event-centric knowledge infusion can introduce external knowledge into model better. The reason is that semantic information in events is clear and lite, and integrating knowledge into events can effectively alleviate effect of noise information. w/o EKI denotes RoBERTa without introducing any external knowledge, and the results show that f1 score and accuracy were only 78.06\% and 78.12\%, which decreased by 1.11\% and 1.08\%, respectively. This shows that it is necessary to inject external knowledge into PTM to improve model's reasoning in scientific texts.

\subsection{Impact of Event-Level Knowledge Infusion}
\begin{table}[t]\small
\setlength\tabcolsep{4.8pt}
\centering
\begin{tabular}{lcccccc}
\hline
\bf Method
& \multicolumn{1}{c}{\bf C} &  \multicolumn{1}{c}{\bf R} &  \multicolumn{1}{c}{\bf E}
&  \multicolumn{1}{c}{\bf N}&  \multicolumn{1}{c}{\bf F1}
&  \multicolumn{1}{c}{\bf ACC}\\ \hline
EventNUM=1& 81.63& 75.19& 77.71& 78.31& 78.21& 78.25\\
EventNUM=5& {\textbf{82.34}}& 75.21& 77.94& 78.28
& 78.44& 78.47\\
EventNUM=10 & 81.76 & {\textbf{76.25}} & {\textbf{78.66}} & {\textbf{80.02}} & {\textbf{79.17}} & {\textbf{79.20}} \\ \hline
\end{tabular}
\caption{Different number of events for retrieval.}\label{tab3}
\end{table}
To investigate the impact of event-level knowledge infusion, we set different numbers of events to retrieval triplets. Concretely, We employ three groups of events with different numbers, and the results are shown in table\ref{tab3}. We can find that when the number of events is less, the F1 score and accuracy of the model decrease significantly. When the number of events is less, knowledge injected into PTM is less, so the model lacks sufficient external knowledge to complement desired knowledge, which leads to worse reasoning.

\subsection{Knowledge Forgetting in Language Model}
\begin{table}[t]\small
\centering
\begin{tabular}{lcc}
\hline
\bf Method  & \bf Triplet ACC & \bf Entity ACC \\ \hline
Sent Mask   & 0.15                            & 0.41                            \\ 
Entity Mask & 0.24                            & 0.42                            \\ \hline
\end{tabular}
\caption{\small Accuracy of the predicted entity and the predicted triplets via fine-tuned PTM. 
\textit{Sent Mask} is to mask random tokens in sentences.
\textit{Entity Mask} is to mask random entities or relations of triplets in sentences.}\label{tab4}
\end{table}
To investigate whether a fine-tuned PTM can memorize and analyze all knowledge in training. We set two mask methods to fine-tune PTM, i.e., Sent Mask and Entity Mask, and results are shown in Table \ref{tab4}. From results, the low ACC shows that the PTM can not accurately learn the scientific knowledge in the sentence after training and occur knowledge forgetting, which also reflects the effectiveness of injecting scientific knowledge from SKG built on the training set.

\subsection{Human Evaluation}
\begin{table}[t]\small
\centering
\begin{tabular}{lcc}
\hline
\bf Method         & \bf Entity ACC                & \bf Relation ACC \\ \hline
SKG &  75.2 &  83.1  \\ \hline
\end{tabular}
\caption{Accuracy on entities and relations in our SKG.}
\label{tab5}
\end{table}
Table \ref{tab5} shows the accuracy of human evaluation for SKG. 
We select 500 triplets randomly selected from SKG, and the accuracy of these 500 triplets was manually measured. We measure the reliability of KG from two views, i.e., entity and relation. Through results on entities and relations, we can see that the quality of our unsupervised constructed knowledge graph is great.

\section{Conclusion}
In this work, we built a scientific knowledge graph in an unsupervised manner. Moreover, we propose event-centric knowledge infusion (EKI) to integrate external knowledge into pre-trained language models. Specifically, we split sentences into multiple events and use them as queries to retrieve triplets in SKG. Moreover, we integrate retrieved knowledge from the built knowledge graph into PTM to help model reasoning at the event level. Experimental results show that our proposed approach achieves state-of-the-art performance on SciNLI tasks, which demonstrates the effectiveness of our method.

\bibliographystyle{IEEEbib}
\bibliography{ref}

\end{document}